# Denoising and compression in wavelet domain via projection onto approximation coefficients

Mario Mastriani

*Abstract*—We describe a new filtering approach in the wavelet domain for image denoising and compression, based on the projections of details subbands coefficients (resultants of the splitting procedure, typical in wavelet domain) onto the approximation subband coefficients (much less noisy). The new algorithm is called Projection Onto Approximation Coefficients (POAC). As a result of this approach, only the approximation subband coefficients and three scalars are stored and/or transmitted to the channel. Besides, with the elimination of the details subbands coefficients, we obtain a bigger compression rate. Experimental results demonstrate that our approach compares favorably to more typical methods of denoising and compression in wavelet domain.

## I. Introduction

An image is affected by noise in its acquisition and processing. The denoising techniques are used to remove the additive noise while retaining as much as possible the important image features. In the recent years there has been an important amount of research on wavelet thresholding and threshold selection for images denoising [1]-[51], because wavelet provides an appropriate basis for separating noisy signal from the image signal. The motivation is that as the wavelet transform is good at energy compaction, the small coefficients are more likely due to noise and large coefficient due to important signal features [1]-[3]. These small coefficients can be thresholded without affecting the significant features of the image.

In fact, the thresholding technique is the last approach based on wavelet theory to provide an enhanced approach for eliminating such noise source [4], [5] and ensure better image quality [6], [7]. Thresholding is a simple non-linear technique, which operates on one wavelet coefficient at a time. In its basic form, each coefficient is thresholded by comparing against threshold, if the coefficient is smaller than threshold, set to zero; otherwise it is kept or modified. Replacing the small noisy coefficients by zero and inverse wavelet transform on the result may lead to reconstruction with the essential signal characteristics and with less noise. Since the work of Donoho & Johnstone [3], there has been much research on finding thresholds, however few are specifically designed for images [14]-[51].

Unfortunately, this technique has the following disadvantages:
1) it depends on the correct election of the type of thresholding, e.g., OracleShrink, VisuShrink (soft-thresholding, hard-thresholding, and semi-soft-thresholding), Sure Shrink, Bayesian soft thresholding, Bayesian MMSE estimation, Thresholding Neural Network (TNN), due to Zhang, NormalShrink, , etc. [1]-[5], [8]-[38]
2) it depends on the correct estimation of the threshold which is arguably the most important design parameter,
3) it doesn't have a fine adjustment of the threshold after their calculation,
4) it should be applied at each level of decomposition, needing several levels, and
5) the specific distributions of the signal and noise may not be well matched at different scales.

Therefore, a new method without these constraints will represent an upgrade. On the other hand, similar considerations should be kept in mind regarding the problem of image compression based on wavelet thresholding.

The Bidimensional Discrete Wavelet Transform and the method to reduce noise and to compress by wavelet thresholding is outlined in Section II. The new approach as denoiser and compression tools in wavelet domain is outlined in Section III. In Section IV, we discuss briefly the more appropriate metrics for denoising and compression. In Section V, the experimental results using the proposed algorithm are presented. Finally, Section VI provides a conclusion of the paper.

## II. Bidimensional Discrete Wavelet Transform

The Bidimensional Discrete Wavelet Transform (DWT-2D) [6]-[7], [12]-[51] corresponds to multiresolution approximation expressions. In practice, mutiresolution analysis is carried out using 4 channel filter banks (for each level of decomposition) composed of a low-pass and a high-pass filter and each filter bank is then sampled at a half rate (1/2 down sampling) of the previous frequency. By repeating this procedure, it is possible to obtain wavelet transform of any order. The down sampling procedure keeps the scaling parameter constant

(equal to ½) throughout successive wavelet transforms so that is benefits for simple computer implementation. In the case of an image, the filtering is implemented in a separable way be filtering the lines and columns.

Note that [6], [7] the DWT of an image consists of four frequency channels for each level of decomposition. For example, for *i*-level of decomposition we have:

$LL_{n,i}$: Noisy Coefficients of Approximation.
$LH_{n,i}$: Noisy Coefficients of Vertical Detail,
$HL_{n,i}$: Noisy Coefficients of Horizontal Detail, and
$HH_{n,i}$: Noisy Coefficients of Diagonal Detail.

The LL part at each scale is decomposed recursively, as illustrated in Fig. 1 [6], [7].

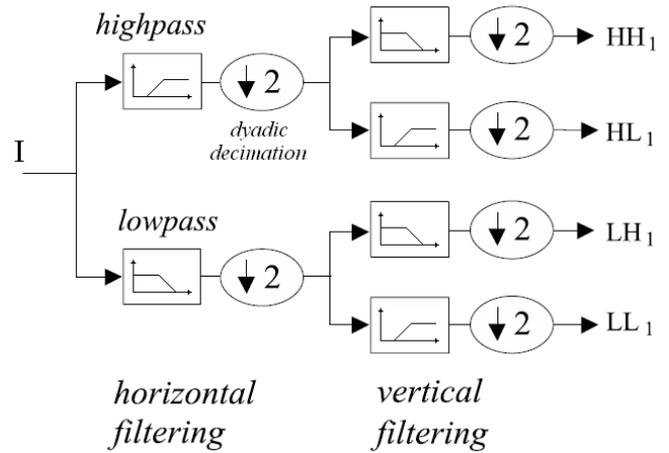

Fig. 2 Two dimensional DWT. A decomposition step.
Usual splitting of the subbands.

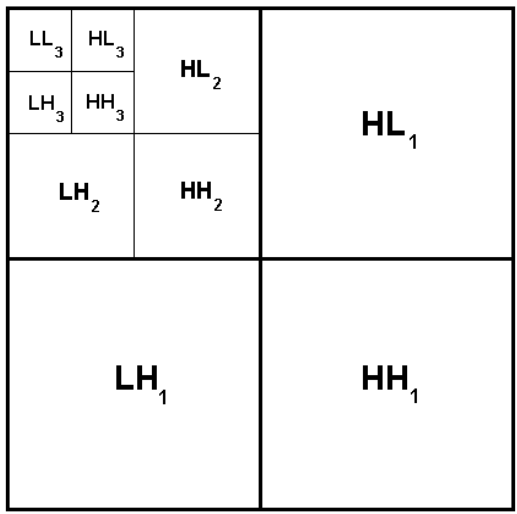

Fig. 1 Data preparation of the image. Recursive decomposition of LL parts.

To achieve space-scale adaptive noise reduction, we need to prepare the 1-D coefficient data stream which contains the space-scale information of 2-D images. This is somewhat similar to the "zigzag" arrangement of the DCT (Discrete Cosine Transform) coefficients in image coding applications [42]. In this data preparation step, the DWT-2D coefficients are rearranged as a 1-D coefficient series in spatial order so that the adjacent samples represent the same local areas in the original image [44].

Figure 2 shows the interior of the DWT-2D with the four subbands of the transformed image [51], which will be used in Fig.3. Each output of Fig. 2 represents a subband of splitting process of the 2-D coefficient matrix corresponding to Fig. 1.

### A. Wavelet Noise Thresholding

The wavelet coefficients calculated by a wavelet transform represent change in the image at a particular resolution. By looking at the image in various resolutions it should be possible to filter out noise, at least in theory. However, the definition of noise is a difficult one. In fact, "one person's noise is another's signal". In part this depends on the resolution one is looking at. One algorithm to remove Gaussian white noise is summarized by D.L. Donoho and I.M. Johnstone [2], [3], and synthesized in Fig. 3.

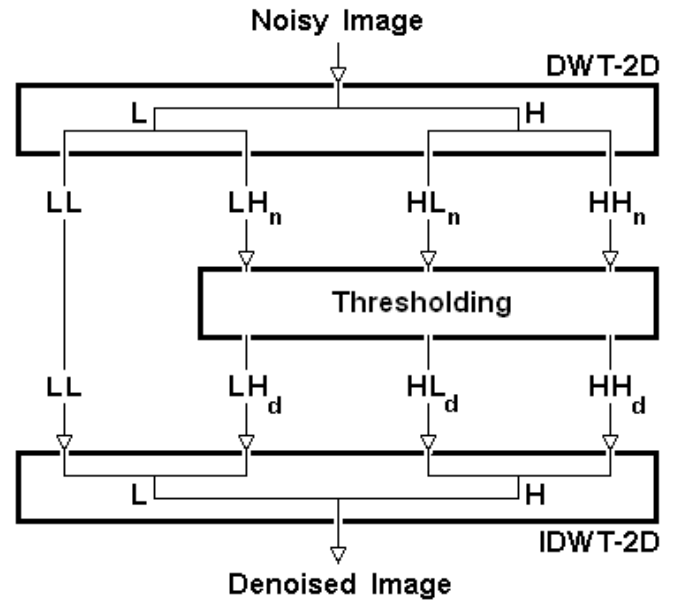

Fig. 3 Thresholding Techniques

The algorithm is:
1) Calculate a wavelet transform and order the coefficients by increasing frequency. This will result in an array containing the image average plus a set of coefficients of length 1, 2, 4, 8, etc. The noise threshold will be calculated on the highest frequency coefficient spectrum (this is the largest spectrum).
2) Calculate the *median absolute deviation* (mad) on the largest coefficient spectrum. The median is calculated from the absolute value of the coefficients. The equation for the median absolute deviation is shown below:

$$\delta_{mad} = \frac{median(|C_{n,i}|)}{0.6745} \qquad (1)$$

where $C_{n,i}$ may be $LH_{n,i}$, $HL_{n,i}$, or $HH_{n,i}$ for *i*-level of decomposition. The factor 0.6745 in the denominator rescales the numerator so that $\delta_{mad}$ is also a suitable estimator for the standard deviation for Gaussian white noise [5], [42], [44].

3) For calculating the noise threshold λ we have used a modified version of the equation that has been discussed in papers by D.L. Donoho and I.M. Johnstone. The equation is:

$$\lambda = \delta_{mad}\sqrt{2log[N]} \qquad (2)$$

where N is the number of pixels in the subimage, i.e., HL, LH or HH.

4) Apply a thresholding algorithm to the coefficients. There are two popular versions:

4.1. Hard thresholding. Hard thresholding sets any coefficient less than or equal to the threshold to zero, see Fig. 4(a).

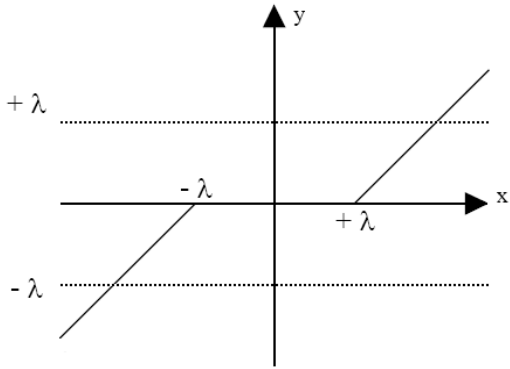

Fig. 4(a) Soft-Thresholfing

where *x* may be $LH_{n,i}$, $HL_{n,i}$, or $HH_{n,i}$, *y* may be $HH_{d,i}$ : Denoised Coefficients of Diagonal Detail,
$HL_{d,i}$ : Denoised Coefficients of Horizontal Detail,
$LH_{d,i}$ : Denoised Coefficients of Vertical Detail,
for *i*-level of decomposition.

The respective code is:

```
for row = 1:N^1/2
  for column = 1:N^1/2
    if |C_n,i[row][column]| <= λ,
      C_n,i[row][column] = 0.0;
    end
  end
end
```

4.2. Soft thresholding. Soft thresholding sets any coefficient less than or equal to the threshold to zero, see Fig. 4(b). The threshold is subtracted from any coefficient that is greater than the threshold. This moves the image coefficients toward zero.

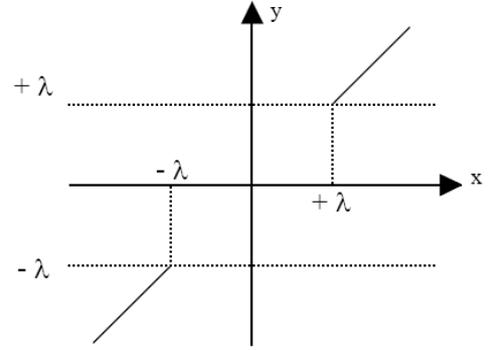

Fig. 4(b): Hard-Thresholfing

The respective code is:

```
for row = 1:N^1/2
  for column = 1:N^1/2
    if |C_n,i[row][column]| <= λ,
      C_n,i[row][column] = 0.0;
    else
      C_n,i[row][column] = C_n,i[row][column] - λ;
    end
  end
end
```

### III. PROJECTION ONTO APPROXIMATION COEFFICIENTS

As a natural consequence of Projection Onto Span Algorithm (POSA), which was introduced by Mastriani [51], the POAC is based on the Orthogonality Principle too [52]. [53].

#### A. Denoising via POAC inside wavelet domain

In this section, the denoising of an image corrupted by white Gaussian noise will be considered, i.e.,

$$I_n = I + n \qquad (3)$$

where *n* is independent Gaussian noise. We observe $I_n$ (a noisy image) and wish to estimate the desired image *I* as accurately as possible according to some criteria.

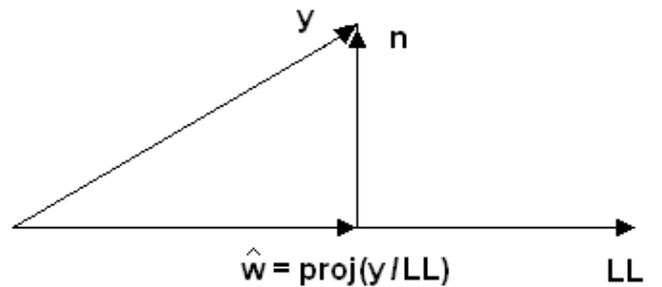

Fig. 5: Orthogonality Principle inside wavelet domain

Inside wavelet domain, if we use an orthogonal wavelet transform, the problem can be formulated as

$$y = \hat{w} + n \tag{4}$$

where *y* noisy wavelet coefficient (LH, HL and HH), $\hat{w}$ true coefficient, and *n* noise, which is independent Gaussian. This is a classical problem in estimation theory [52]. Our aim is to estimate from the noisy observation. A estimator based on the orthogonality principle will be used for this purpose [52], [53]. Such estimators have been widely advocated for image restoration and reconstruction problems [51], [54]. In this particular case, and based on Fig.5, we have

$$\hat{w} = proj\,(y\,/\,LL) \tag{5}$$

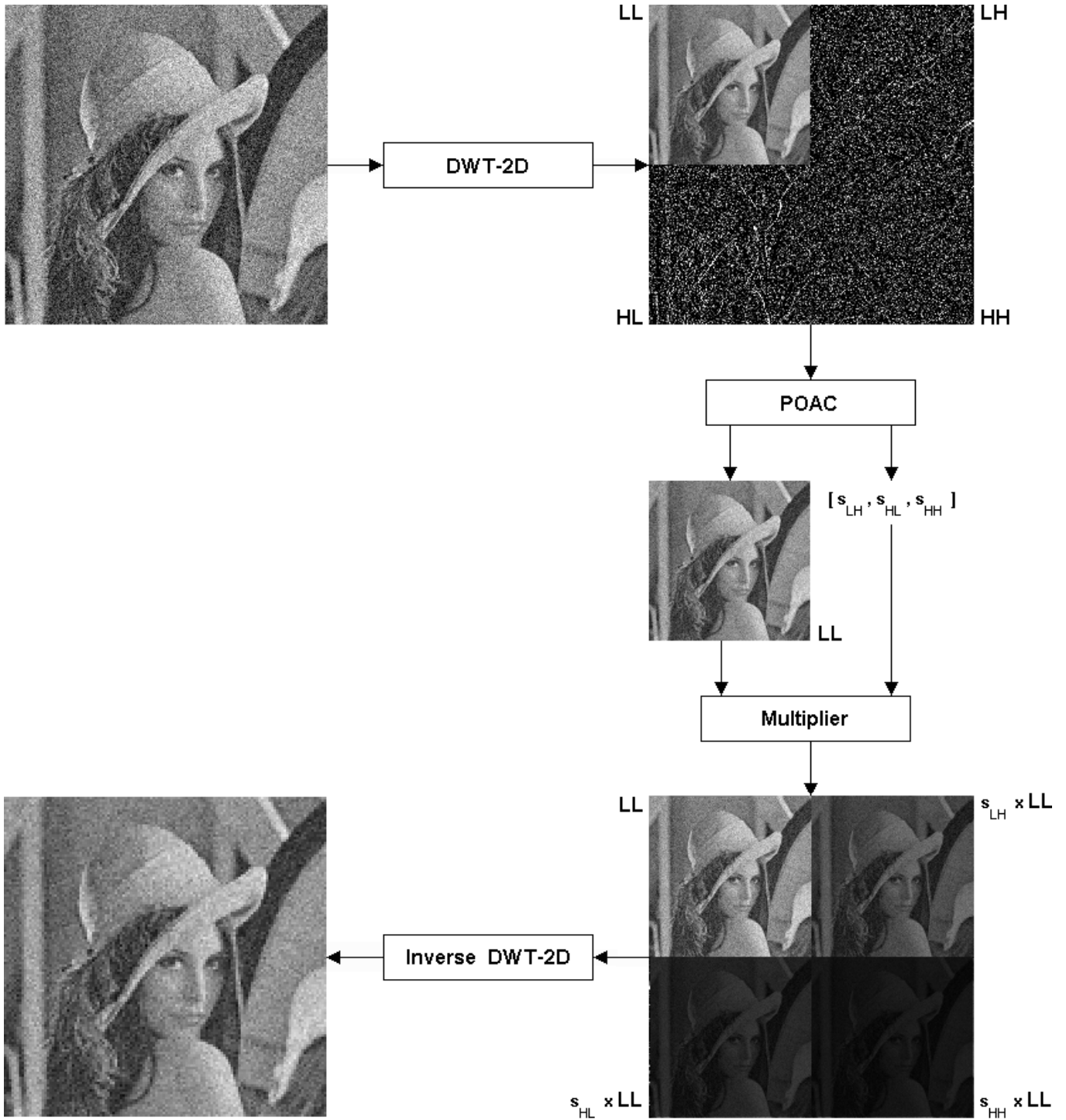

Fig. 6: POAC algorithm as denoiser

That is to say, $\hat{w}$ *is the projection of y* (LH, HL and HH) *onto LL.* Therefore,

$$\hat{w} = \frac{trace(LL\ y^T)}{trace(LL\ LL^T)} LL \quad (6)$$

arising three possibilities, i.e.,

$$\hat{w} = s_{LH}\ LL \quad (7.a)$$

$$\hat{w} = s_{HL}\ LL \quad (7.b)$$

$$\hat{w} = s_{HH}\ LL \quad (7.c)$$

where

$$s_{LH} = \frac{trace(LL\ LH^T)}{trace(LL\ LL^T)} \quad (8.a)$$

$$s_{HL} = \frac{trace(LL\ HL^T)}{trace(LL\ LL^T)} \quad (8.b)$$

$$s_{HH} = \frac{trace(LL\ HH^T)}{trace(LL\ LL^T)} \quad (8.c)$$

That is to say, they are three scalars that arising as a consequence of POAC application inside wavelet domain (see Fig.6). This allows generate three new denoised detail coefficient matrices, uncorrelated regarding the noise and correlated with the approximation coefficient matrix LL, the less noisy one of all.

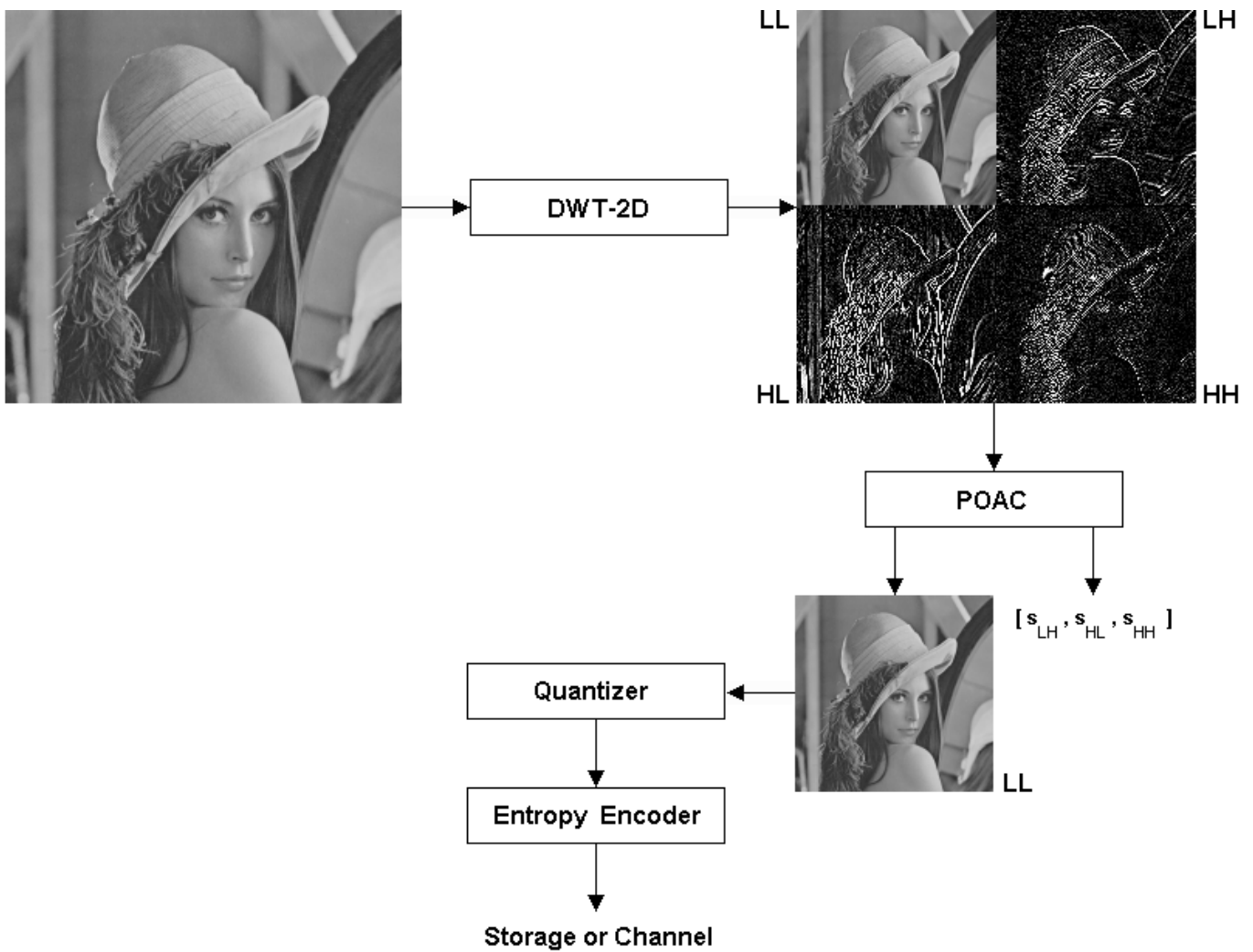

Fig. 7(a): POAC algorithm as compressor. ENCODER

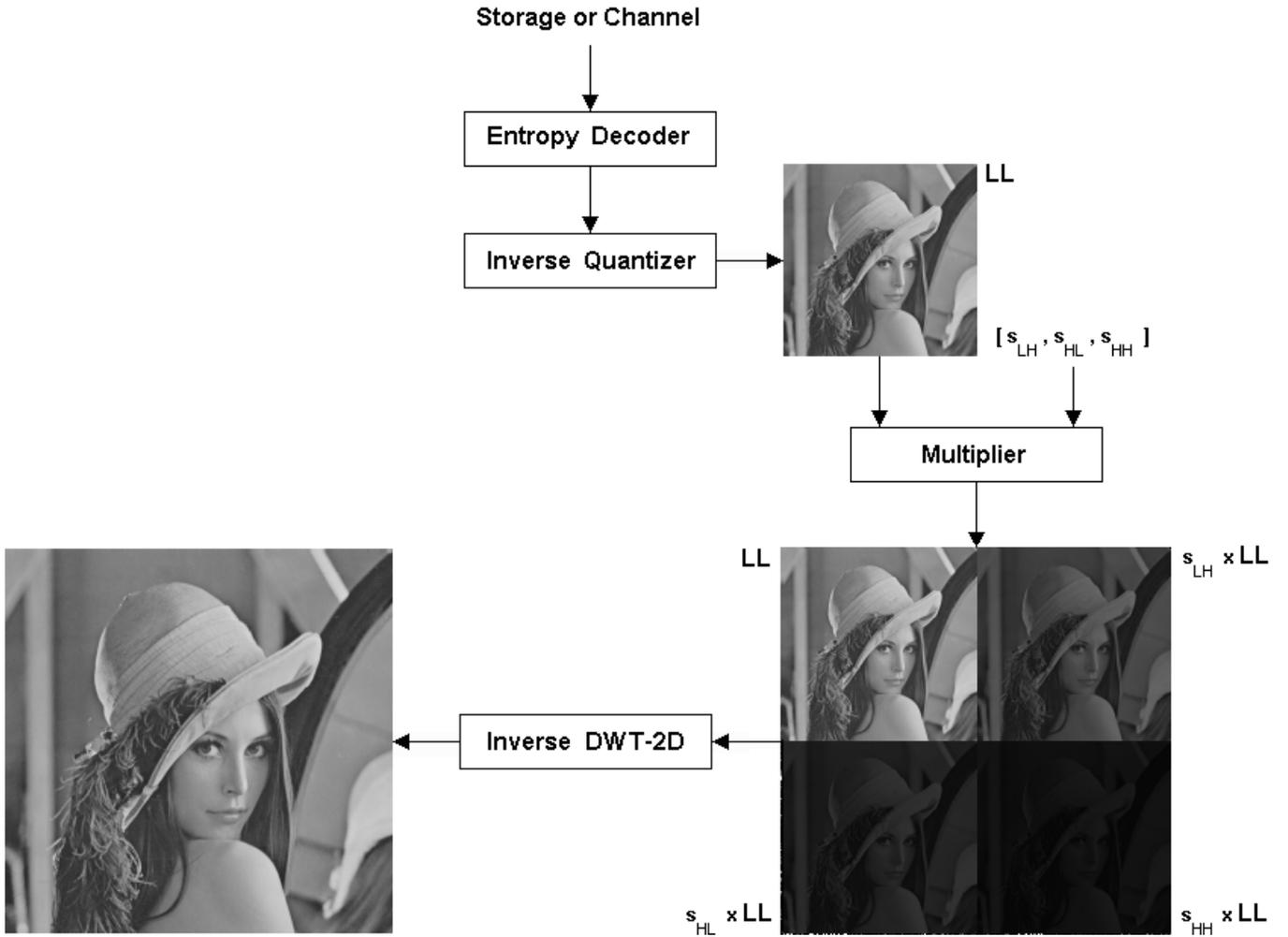

Fig. 7(b): POAC algorithm as compressor. DECODER

### B. Compression thanks to POAC inside wavelet domain

As we could see in the previous section, the input of POAC inside wavelet domain, are the four subbands, i.e., LL, LH, HL and HH, while its output is the approximation subband LL plus three scalars $s_{LH}$, $s_{HL}$ and $s_{HH}$. This intrinsically, represents a compression approach with a compression rate of 4:1, approximately.

The Figures 7(a) and 7(b) represents the encoder and decoder architecture for compression thanks POAC inside wavelet domain.

## IV. METRICS

### A. Data Compression Ratio (CR)

Data compression ratio, also known as compression power, is a computer-science term used to quantify the reduction in data-representation size produced by a data compression algorithm. The data compression ratio is analogous to the physical compression ratio used to measure physical compression of substances, and is defined in the same way, as the ratio between the *uncompressed size* and the *compressed size* [54]:

$$CR = \frac{Uncompressed\ Size}{Compressed\ Size} \qquad (9)$$

Thus a representation that compresses a 10MB file to 2MB has a compression ratio of 10/2 = 5, often notated as an explicit ratio, 5:1 (read "five to one"), or as an implicit ratio, 5X. Note that this formulation applies equally for compression, where the uncompressed size is that of the original; and for decompression, where the uncompressed size is that of the reproduction.

### B. Percent Space Savings (PSS)

Sometimes the space savings is given instead, which is defined as the reduction in size relative to the uncompressed

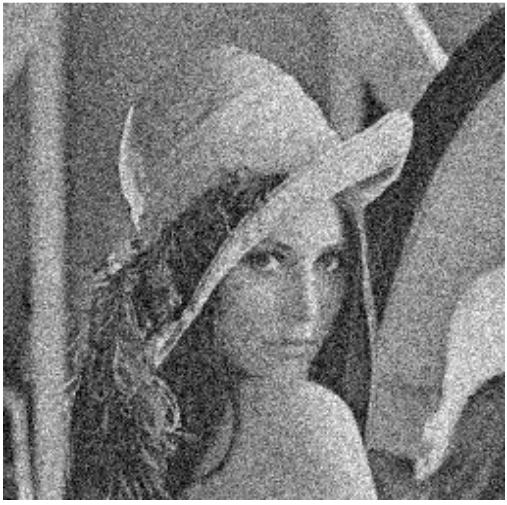
Noisy

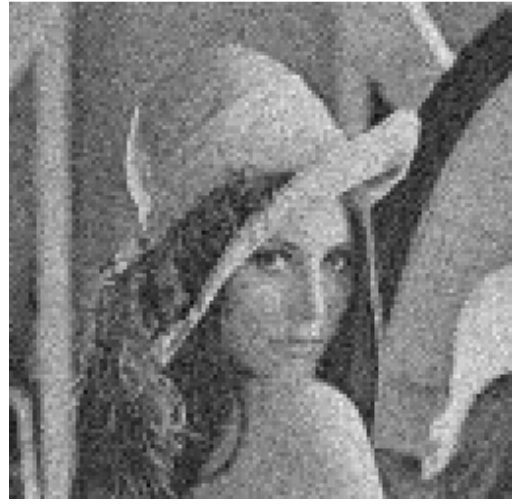
Soft-thresholding (ST)

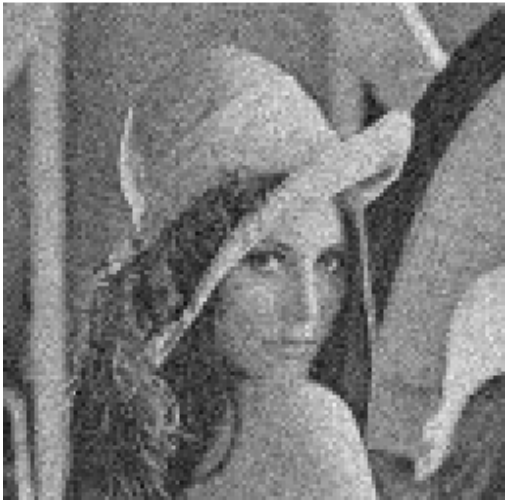
Hard-thresholding (HT)

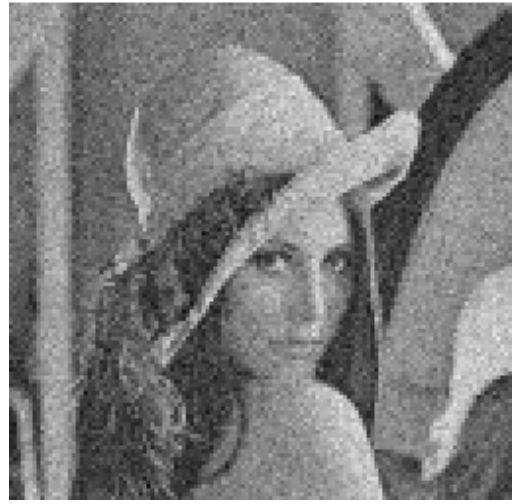
POAC

Fig. 8: Noisy and filtered images.

size:

$$PSS = (1 - \frac{1}{CR}) * 100\% \qquad (10)$$

Thus a representation that compresses 10MB file to 2MB would yield a space savings of 1-2/10 = 0.8, often notated as a percentage, 80%.

### C. Peak Signal-To-Noise Ratio (PSNR)

The phrase peak signal-to-noise ratio, often abbreviated PSNR, is an engineering term for the ratio between the maximum possible power of a signal and the power of corrupting noise that affects the fidelity of its representation. Because many signals have a very wide dynamic range, PSNR is usually expressed in terms of the logarithmic decibel scale.

The PSNR is most commonly used as a measure of quality of reconstruction in image compression etc [54]. It is most easily defined via the mean squared error (MSE) which for two $NR \times NC$ (rows-by-columns) monochrome images $I$ and $I_d$, where the second one of the images is considered a denoised approximation of the other is defined as:

$$MSE = \frac{1}{NRxNC} \sum_{nr=0}^{NR-1} \sum_{nc=0}^{NC-1} \|I(nr,nc) - I_d(nr,nc)\|^2 \qquad (11)$$

The PSNR is defined as [ ]:

$$PSNR = 10\log_{10}(\frac{MAX_I^2}{MSE}) = 20\log_{10}(\frac{MAX_I}{\sqrt{MSE}}) \qquad (12)$$

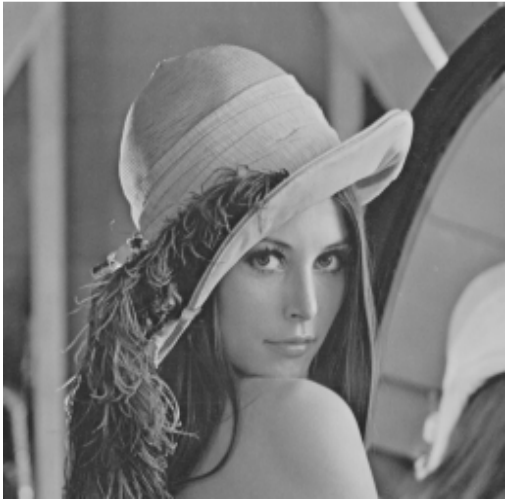
Original

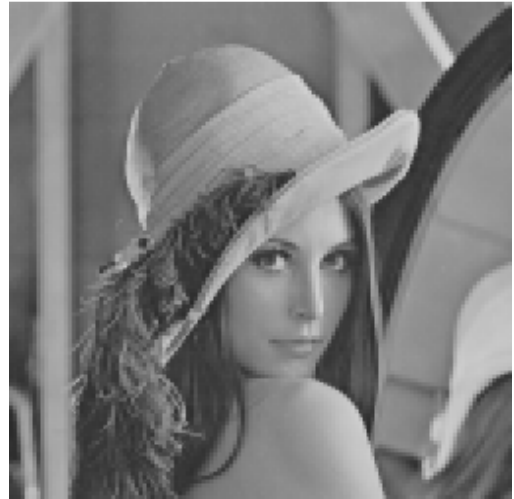
Soft-thresholding (ST)

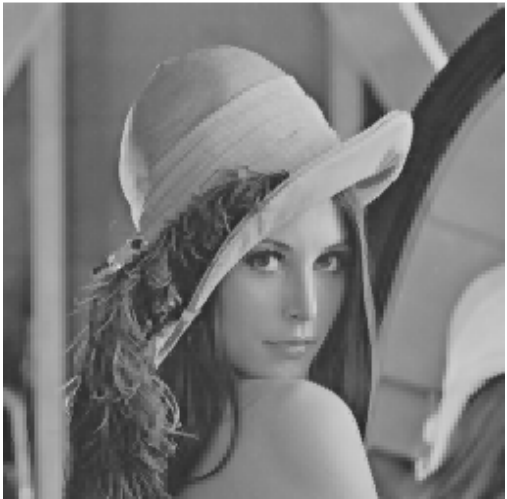
Hard-thresholding (HT)

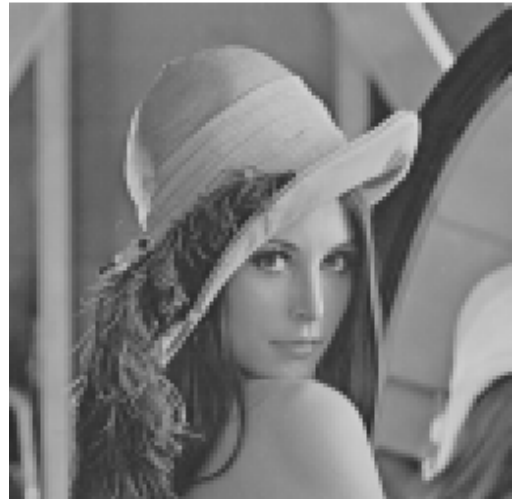
POAC

Fig. 9: Original and compressed images.

Here, $MAX_i$ is the maximum pixel value of the image. When the pixels are represented using 8 bits per sample, this is 255. More generally, when samples are represented using linear pulse code modulation (PCM) with B bits per sample, maximum possible value of $MAX_i$ is $2^B-1$.

For color images with three red-green-blue (RGB) values per pixel, the definition of PSNR is the same except the MSE is the sum over all squared value differences divided by image size and by three [54].

Typical values for the PSNR in lossy image and video compression are between 30 and 50 dB, where higher is better.

## V. EXPERIMENTAL RESULTS

The simulations demonstrate that the POAC technique improves the noise reduction and compression performances in wavelet domain to the maximum.

Here, we present a set of experimental results using one typical image. Such images were converted to bitmap file format for their treatment [54]. Figure 8 shows the noisy (Gaussian white noise, with mean value = 0, and standard deviation = 0.01) and filtered images, with 256-by-256 (pixel) by 256 (gray levels) bitmap matrix. Table I summarizes the assessment parameters vs. filtering techniques for Fig.8, where ST means Soft-Thresholding and HT means Hard-Thresholding. On the other hand, Fig.9 shows the original and compressed /decompressed images via ST, HT and POAC techniques. Table II summarizes the assessment parameters vs. compressed techniques for Fig.9. The quality is similar with very different CR and PSS.

TABLE I
ORIGINAL VS DENOISED IMAGES

| METRIC | ST | HT | POAC |
|---|---|---|---|
| MSE | 229.2780 | 229.2803 | 228.6764 |
| PSNR | 24.5272 | 24.5271 | 24.5386 |

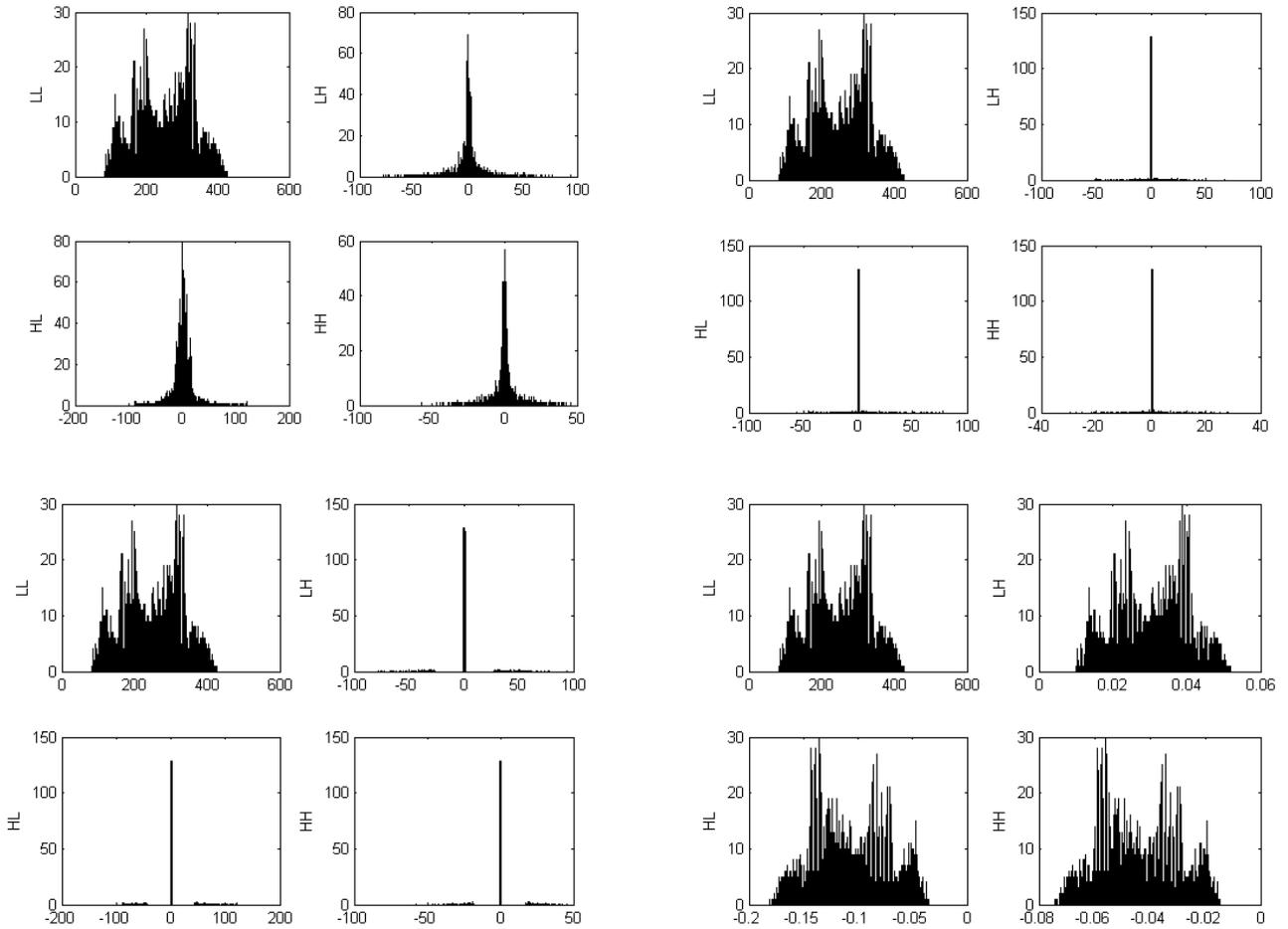

Fig. 10: Histograms of wavelet coefficients: (a) before thresholding, (b) after soft-thresholding, (c) after hard-thresholding, and (d) after POAC

TABLE II
ORIGINAL VS DECOMPRESSED IMAGES

| METRIC | ST | HT | POAC |
|---|---|---|---|
| CR | 3.065 | 3.044 | 5.055 |
| PSS [%] | 67.37 | 67.15 | 80.21 |
| MSE | 48.3938 | 34.9821 | 63.6266 |
| PSNR | 31.2829 | 32.6923 | 30.0944 |

On the other hand, Fig.10 shows the histograms of approximation (LL) and detail (LH, HL and HH) wavelet coefficients before and after the thresholding techniques. Observe, the damage caused for the pruning of ST and HT techniques, and the histogram affinity between LL (less noisy) and wavelet coefficients (LH, HL and HH) after POAC technique; considering that the histogram depends on the noisy presence in the wavelet coefficients.

Wavelet basis employed in the experiments were Daubechies 1, 2 and 4, with only one level of decomposition.

Finally, all techniques (denoising and compression) were implemented in MATLAB® (Mathworks, Natick, MA) on a PC with an Athlon (2.4 GHz) processor.

## VI. CONCLUSION

In this paper we have developed a *Projection Onto Approximation Coefficients* technique for image filtering and compression inside wavelet domain. The simulations show that the POAC have better performance than the most commonly used thresholding technique for compression and denoising (for the studied benchmark parameters) which include Soft-Thresholding and Hard-Thresholding.

Besides, the novel demonstrated to be efficient to remove multiplied noise, and all uncle of noise in the undecimated wavelet domain. Finally, cleaner images suggest potential improvements for classification and recognition.

ACKNOWLEDGMENT

M. Mastriani thanks Prof. Marta Mejail, director of Laboratorio de Visión Robótica y Procesamiento de Imágenes del Departamento de Computación de la Facultad de Ciencias Exactas y Naturales de la Universidad de Buenos Aires, for her tremendous help and support.

REFERENCES

[1] D. L. Donoho, "De-noising by soft-thresholding," IEEE Trans. Inform. Theory, vol. 41, no. 3, pp. 613-627, 1995.

[2] D. L. Donoho, and I. M. Johnstone, "Adapting to unknown smoothness via wavelet shrinkage," Journal of the American Statistical Assoc., vol. 90, no. 432, pp. 1200-1224., 1995.

[3] D. L. Donoho, and I. M. Johnstone, "Ideal spatial adaptation by wavelet shrinkage," Biometrika, 81, 425-455, 1994.

[4] I. Daubechies. Ten Lectures on Wavelets, SIAM, Philadelphia, PA. 1992.

[5] I. Daubechies, "Different Perspectives on Wavelets," in *Proceedings of Symposia in Applied Mathematics*, vol. 47, American Mathematical Society, USA, 1993.

[6] S. Mallat, "A theory for multiresolution signal decomposition: The wavelet representation," *IEEE Trans. Pattern Anal. Machine Intell.*, vol. 11, pp. 674–693, July 1989.

[7] S. G. Mallat, "Multiresolution approximations and wavelet orthonormal bases of L2 (R)," *Transactions of the American Mathematical Society*, 315(1), pp.69-87, 1989a.

[8] X.-P. Zhang, and M. Desai, "Nonlinear adaptive noise suppression based on wavelet transform," Proceedings of the ICASSP98, vol. 3, pp. 1589-1592, Seattle, 1998.

[9] X.-P. Zhang, "Thresholding Neural Network for Adaptive Noise reduction," IEEE Transactions on Neural Networks, vol.12, no. 3, pp.567-584, 2001.

[10] X.-P. Zhang, and M. Desai, "Adaptive Denoising Based On SURE Risk," IEEE Signal Proc. Letters, vol.5, no. 10, 1998.

[11] X.-P. Zhang and Z.Q. Luo, "A new time-scale adaptive denoising method based on wavelet shrinkage," *in Proceedings of the ICASSP99*, Phoenix, AZ., March 15-19, 1999.

[12] M. Lang, H. Guo, J. Odegard, C. Burrus, and R. Wells, "Noise reduction using an undecimated discrete wavelet transform," IEEE Signal Proc. Letters, vol. 3, no. 1, pp. 10-12, 1996.

[13] H. Chipman, E. Kolaczyk, and R. McCulloch, "Adaptive Bayesian wavelet shrinkage," *J. Amer. Statist. Assoc.*, vol. 92, pp. 1413–1421, 1997.

[14] S. G. Chang, B. Yu, and M. Vetterli, "Spatially adaptive wavelet thresholding with context modeling for image denoising," *IEEE Trans. Image Processing*, vol. 9, pp. 1522–1531, Sept. 2000.

[15] S. G. Chang, B. Yu, and M. Vetterli, "Adaptive wavelet thresholding for image denoising and compression," *IEEE Trans. Image Processing*, vol. 9, pp. 1532–1546, Sept. 2000.

[16] S. G. Chang and M. Vetterli, "Spatial adaptive wavelet thresholding for image denoising," in *Proc. ICIP*, vol. 1, 1997, pp. 374–377.

[17] M. S. Crouse, R. D. Nowak, and R. G. Baraniuk, "Wavelet-based statistical signal processing using hidden Markov models," *IEEE Trans.Signal Processing*, vol. 46, pp. 886–902, Apr. 1998.

[18] M. Malfait and D. Roose, "Wavelet-based image denoising using a Markov random field *a priori* model," *IEEE Trans. Image Processing*, vol. 6, pp. 549–565, Apr. 1997.

[19] M. K. Mihcak, I. Kozintsev, K. Ramchandran, and P. Moulin, "Low complexity image denoising based on statistical modeling of wavelet coefficients," *IEEE Trans. Signal Processing Lett.*, vol. 6, pp. 300–303, Dec. 1999.

[20] E. P. Simoncelli, "Bayesian denoising of visual images in the wavelet domain," in *Bayesian Inference in Wavelet Based Models*. New York: Springer-Verlag, 1999, pp. 291–308.

[21] E. Simoncelli and E. Adelson, "Noise removal via Bayesian wavelet coring," in *Proc. ICIP*, vol. 1, 1996, pp. 379–382.

[22] M. Belge, M. E. Kilmer, and E. L. Miller, "Wavelet domain image restoration with adaptive edge-preserving regularization," *IEEE Trans. Image Processing*, vol. 9, pp. 597–608, Apr. 2000.

[23] J. Liu and P. Moulin, "Information-theoretic analysis of interscale and intrascale dependencies between image wavelet coefficients," *IEEE Trans. Image Processing*, vol. 10, pp. 1647–1658, Nov. 2000.

[24] H. Guo, J. E. Odegard, M. Lang, R. A. Gopinath, I. Selesnick, and C. S. Burrus, "Speckle reduction via wavelet shrinkage with application to SAR based ATD/R," Technical Report CML TR94-02, CML, Rice University, Houston, 1994.

[25] R.R. Coifman, and D.L. Donoho, *Translation-invariant de-noising*. A. Antoniadis & G. Oppenheim (eds), Lecture Notes in Statistics, vol. 103. Springer-Verlag, pp 125-150, 1995.

[26] M. Misiti, Y. Misiti, G. Oppenheim, and J.M. Poggi. (2001, June). Wavelet Toolbox, for use with MATLAB®, User's guide, version 2.1. [Online]. Available: http://www.rrz.uni-hamburg.de/RRZ/Software/Matlab/Dokumentation/help/pdf_doc/wavelet/wavelet_ug.pdf

[27] C.S. Burrus, R.A. Gopinath, and H. Guo, *Introduction to Wavelets and Wavelet Transforms: A Primer*, Prentice Hall, New Jersey, 1998.

[28] B.B. Hubbard, *The World According to Wavelets: The Story of a Mathematical Technique in the Making*, A. K. Peter Wellesley, Massachusetts, 1996.

[29] A. Grossman and J. Morlet, "Decomposition of Hardy Functions into Square Integrable Wavelets of Constant Shape," *SIAM J. App Math*, 15: pp.723-736, 1984.

[30] C. Valens. (2004). A really friendly guide to wavelets. [Online]. Available: http://perso.wanadoo.fr/polyvalens/clemens/wavelets/wavelets.html

[31] G. Kaiser, *A Friendly Guide To Wavelets*, Boston: Birkhauser, 1994.

[32] J.S. Walker, *A Primer on Wavelets and their Scientific Applications*, Chapman & Hall/CRC, New York, 1999.

[33] E. J. Stollnitz, T. D. DeRose, and D. H. Salesin, *Wavelets for Computer Graphics: Theory and Applications*, Morgan Kaufmann Publishers, San Francisco, 1996.

[34] J. Shen and G. Strang, "The zeros of the Daubechies polynomials," in *Proc. American Mathematical Society,* 1996.

[35] R. Yu, A.R. Allen, and J. Watson, *An optimal wavelet thresholding for speckle noise reduction*, in Summer School on Wavelets: Papers, Publisher: Silesian Technical University (Gliwice, Poland), pp77-81, 1996.

[36] H.Y. Gao, and A.G. Bruce, "WaveShrink with firm shrinkage," *Statistica Sinica*, 7, 855-874, 1997.

[37] L. Gagnon, and F.D. Smaili, "Speckle noise reduction of air-borne SAR images with Symmetric Daubechies Wavelets," in *SPIE Proc. #2759*, pp. 1424, 1996.

[38] M.S. Crouse, R. Nowak, and R. Baraniuk, "Wavelet-based statistical signal processing using hidden Markov models," *IEEE Trans. Signal Processing*, vol 46, no.4, pp.886-902, 1998.

[39] M. Mastriani and A. Giraldez, "*Smoothing of coefficients in wavelet domain for speckle reduction in Synthetic Aperture Radar images*," ICGST International Journal on Graphics, Vision and Image Processing (GVIP), Volume 6, pp. 1-8, 2005. [Online]. Available: http://www.icgst.com/gvip/v6/P1150517003.pdf

[40] M. Mastriani y A. Giraldez, "*Despeckling of SAR images in wavelet domain*," GIS Development Magazine, Sept. 2005, Vol. 9, Issue 9, pp.38-40. [Online]. Available: http://www.gisdevelopment.net/magazine/years/2005/sep/wavelet_1.htm

[41] M. Mastriani y A. Giraldez, "*Microarrays denoising via smoothing of coefficients in wavelet domain*," WSEAS Transactions on Biology and Biomedicine, 2005. [Online]. Available: http://www.wseas.org/online

[42] M. Mastriani y A. Giraldez, "*Fuzzy thresholding in wavelet domain for speckle reduction in Synthetic Aperture Radar images*," ICGST International on Journal of Artificial Intelligence and Machine Learning, Volume 5, 2005. [Online]. Available: http://www.icgst.com/aiml/v3/index.html

[43] M. Mastriani, "*Denoising based on wavelets and deblurring via self- organizing map for Synthetic Aperture Radar images*," ICGST International on Journal of Artificial Intelligence and Machine Learning, Volume 5, 2005. [Online]. Available: http://www.icgst.com/aiml/v3/index.html

[44] M. Mastriani, "*Systholic Boolean Orthonormalizer Network in Wavelet Domain for Microarray Denoising*," ICGST International Journal on Bioinformatics and Medical Engineering, Volume 5, 2005. [Online]. Available: http://www.icgst.com/bime/v1/bimev1.html

[45] M. Mastriani, "*Denoising based on wavelets and deblurring via self-organizing map for Synthetic Aperture Radar images*," International Journal of Signal Processing, Volume 2, Number 4, pp.226-235, 2005. [Online]. Available: http://www.enformatika.com/ijsp/v2/v2-4-33.pdf

[46] M. Mastriani, "*Systholic Boolean Orthonormalizer Network in Wavelet Domain for Microarray Denoising*," International Journal

of Signal Processing, Volume 2, Number 4, pp.273-284, 2005. [Online]. Available: http://www.enformatika.org/ijsp/v2/v2-4-40.pdf
[47] M. Mastriani y A. Giraldez, "*Microarrays denoising via smoothing of coefficients in wavelet domain*," International Journal of Biomedical Sciences, Volume 1, Number 1, pp.7-14, 2006. [Online]. Available: http://www.enformatika.org/ijbs/v1/v1-1-2.pdf
[48] M. Mastriani y A. Giraldez, "*Kalman' Shrinkage for Wavelet-Based Despeckling of SAR Images*," International Journal of Intelligent Technology, Volume 1, Number 3, pp.190-196, 2006. [Online]. Available: http://www.enformatika.org/jit/v1/v1-3-21.pdf
[49] M. Mastriani y A. Giraldez, "*Neural Shrinkage for Wavelet-Based SAR Despeckling*," International Journal of Intelligent Technology, Volume 1, Number 3, pp.211-222, 2006. [Online]. Available: http://www.enformatika.org/jit/v1/v1-3-24.pdf
[50] M. Mastriani, "*Fuzzy Thresholding in Wavelet Domain for Speckle Reduction in Synthetic Aperture Radar Images*," International Journal of Intelligent Technology, Volume 1, Number 3, pp.252-265, 2006. [Online]. Available: http://www.enformatika.org/jit/v1/v1-3-30.pdf
[51] M. Mastriani, "*New Wavelet-Based Superresolution Algorithm for Speckle Reduction in SAR Images*," International Journal of Computer Science, Volume 1, Number 4, pp.291-298, 2006. [Online]. Available: http://www.enformatika.org/ijcs/v1/v1-4-39.pdf
[52] S. Haykin. Adaptive Filter Theory, Prentice-Hall, Englewood Cliffs, New Jersey, 1986.
[53] Leon S. J., Linear Algebra with Applications, MacMillan, 1990, New York.
[54] A.K. Jain, *Fundamentals of Digital Image Processing,* Englewood Cliffs, New Jersey, 1989.